\documentclass[sigconf]{acmart}

\usepackage{booktabs} 
\usepackage{graphicx}
\usepackage{url}

\setcopyright{rightsretained}

\acmConference[RDSM, CIKM'18]{2nd International Workshop on Rumours and Deception in Social Media (RDSM) 
co-located with International Conference on Information and Knowledge Management}{October 2018}{
 Torino, Italy}
\acmYear{2018}
\copyrightyear{2018}

\begin{document}
\title{Stance Classification for Rumour Analysis in Twitter: \\ Exploiting Affective Information and Conversation Structure}

\author{Endang Wahyu Pamungkas}
\affiliation{%
  \institution{Dipartimento di Informatica, University of Turin}
  \streetaddress{Corso Svizzera, 185}
  \city{Turin}
  \state{Italy}
  \postcode{10149}
}
\email{pamungka@di.unito.it}

\author{Valerio Basile}
\affiliation{%
  \institution{Dipartimento di Informatica, University of Turin}
  \streetaddress{Corso Svizzera, 185}
  \city{Turin}
  \state{Italy}
  \postcode{10149}
}
\email{basile@di.unito.it}

\author{Viviana Patti}
\affiliation{%
  \institution{Dipartimento di Informatica, University of Turin}
  \streetaddress{Corso Svizzera, 185}
  \city{Turin}
  \state{Italy}
  \postcode{10149}
}
\email{patti@di.unito.it}

\begin{abstract}
Analysing how people react to rumours associated with news in social media is an important task to prevent the spreading of misinformation, which is nowadays widely recognized as a dangerous tendency. 
In social media conversations, users show different stances and attitudes towards rumourous stories. Some users take a definite stance, supporting or denying the rumour at issue, while others just comment it, or ask for additional evidence related to the veracity of the rumour. On this line, a new 
shared task has been proposed at SemEval-2017 (Task 8, SubTask A), which is focused on rumour stance classification in English tweets. The goal is predicting user stance towards emerging rumours in Twitter, in terms of supporting, denying, querying, or commenting the original rumour, looking at the conversation threads originated by the rumour. This paper describes a new approach to this task, where the use of conversation-based and affective-based features, covering different facets of affect, has been explored. 
Our classification model outperforms the best-performing systems for stance classification at SemEval-2017 Task 8,
showing the effectiveness of the feature set proposed.
\end{abstract}

%
%



\maketitle

\section{Introduction}
Nowadays, people increasingly tend to use social media like Facebook and Twitter as their primary source of information and news consumption. 
There are several reasons behind this tendency, such as
the simplicity to gather and share the news and the possibility of
staying abreast of the latest news and updated faster than with traditional media.
An important factor is also that 
people can be engaged in conversations on the latest breaking news with their contacts by using these platforms. 
Pew Research Center's newest report\footnote{\url{http://www.journalism.org/2017/09/07/news-use-across-social-media-platforms-2017/}}  shows that two-thirds of U.S. adults gather their news from social media, where Twitter is the most used platform. 
However, 
the absence of a systematic approach to do some form of fact and veracity checking
may also encourage the spread of rumourous stories and 
misinformation 
\cite{procter2013reading}. Indeed, in social media, unverified information can spread very quickly and becomes viral easily, enabling the diffusion of false rumours and fake information. 

Within this scenario, it is crucial to analyse people attitudes towards rumours in social media and to resolve their veracity as soon as possible. 
Several approaches have been proposed to check the rumour veracity in social media 
~\cite{shu2017fake}.
This paper focus on a stance-based analysis of event-related rumours, following the approach proposed at 
SemEval-2017 in the new RumourEval shared task (Task 8, sub-task A)~\cite{derczynski2017semeval}.
In this task English tweets from conversation threads, each associated to a newsworthy event and the rumours around it, are provided as data. The goal is to determine whether a tweet in the thread is supporting, denying, querying, or commenting the original rumour which started the conversation. 
It can be considered a stance classification task, where we have to predict the user's stance towards the rumour from a tweet, in the context of a given thread. This task has been defined as {\em open stance classification task} and is conceived as a key step in rumour resolution, by providing an analysis of people reactions towards an emerging rumour \cite{procter2013reading,zubiaga2016analysing}. The task is also different from detecting stance towards a specific target entity \cite{mohammad2016semeval}.

{\bf Contribution} We describe a novel classification approach, by proposing a new feature matrix, which includes two new groups:  
(a) features exploiting the conversational structure of the dataset \cite{derczynski2017semeval};
(b) affective features relying on the use of a wide range of affective resources capturing different facets of sentiment and other affect-related phenomena.
We were also inspired by the fake news study on Twitter in \cite{vosoughi2018spread}, showing that false stories inspire fear, disgust, and surprise in replies, while true stories inspire anticipation, sadness, joy, and trust. Meanwhile, from a dialogue act perspective, the study of \cite{novielli2013role} found that a relationship exists between the use of an affective lexicon and the communicative intention of an utterance which includes AGREE-ACCEPT (support), REJECT (deny), INFO-REQUEST (question), and OPINION (comment). They exploited several LIWC categories to analyse the role of affective content. 

Our results show that our model outperforms the state of the art on the Semeval-2017 benchmark dataset.
%
Feature analysis highlights the contribution of the different feature groups, and error analysis is shedding some light on the main difficulties and challenges which still need to be addressed.   

{\bf Outline} The paper is organized as follows. Section 2 introduces the SemEval-2017 Task 8. Section 3 describes our approach to deal with open stance classification by exploiting different groups of features. Section 4 describes the evaluation and includes a qualitative error analysis. Finally, Section 5 concludes the paper and points to future
directions.

\section{SemEval-2017 Task 8: RumourEval} \label{sec:dataset}

The SemEval-2017 Task 8 Task A \cite{derczynski2017semeval} has as its main objective to determine the stance of the users in a Twitter thread towards a given rumour, in terms of support, denying, querying or commenting (SDQC) on the original rumour. 
Rumour is defined as a {\em ``circulating story of questionable veracity, which is apparently credible but hard to verify, and produces sufficient skepticism and/or anxiety so as to motivate finding out the actual truth''} \cite{zubiaga2015towards}. 
The task was very timing due to the growing importance of rumour resolution in the breaking news and to the urgency of preventing the spreading of misinformation. 

 {\bf Dataset}\footnote{\url{http://alt.qcri.org/semeval2017/task8/index.php?id=data-and-tools}}   The data for this task are taken from Twitter conversations about news-related rumours collected by \cite{zubiaga2016analysing}. They were annotated using four labels (SDQC): {\em support - S} (when tweet's author support the rumour veracity); {\em deny -D} (when tweet's author denies the rumour veracity); {\em query - Q} (when tweet's author ask for additional information/evidence); {\em comment -C} (when tweet's author just make a comment and does not give important information to asses the rumour veracity). 
The distribution consists of three sets: development, training and test sets, as summarized in Table \ref{data-distribution}, where you can see also the label distribution and the news related to the rumors discussed. 
Training data consist of 297 Twitter conversations and 4,238 tweets in total with related direct and nested replies, where conversations are associated to seven different breaking news. Test data consist of 1049 tweets, where two new rumourous topics were added. 

\begin{table}
\begin{center}
\begin{tabular}{ p{3.1cm} p{0.6cm}p{0.6cm}p{0.6cm}p{0.6cm}  }
 \hline
 \multicolumn{5}{c}{\textbf{Development Data}} \\
 \hline
   \textbf{Rumour} & \textbf{S} & \textbf{D} & \textbf{Q} & \textbf{C}\\
 \hline
 Germanwings & 69 & 11 & 28 & 173\\
 \hline
 \hline
 \multicolumn{5}{c}{\textbf{Training Data}} \\
 \hline
   \textbf{Rumour} & \textbf{S} & \textbf{D} & \textbf{Q} & \textbf{C}\\
 \hline
 Charlie Hebdo & 239 & 58 & 53 & 721\\
 Ebola Essien &  6 & 6 & 1 & 21\\
 Ferguson &  176 & 91 & 99 & 718\\
 Ottawa Shooting &  161 & 76 & 63 & 477\\
 Prince Toronto &  21 & 7 & 11 & 64\\
 Putin Missing &  18 & 6 & 5 & 33\\
 Sydney Siege &  220 & 89 & 98 & 700\\
 \hline
 \textbf{Total} & 841 & 333 & 330 & 2734\\
 \hline
 \hline
 \multicolumn{5}{c}{\textbf{Testing Data}} \\
 \hline
   \textbf{Rumour} & \textbf{S} & \textbf{D} & \textbf{Q} & \textbf{C}\\
 \hline
 Ferguson &  15 & 4 & 17 & 66\\
 Ottawa Shooting &  10 & 2 & 20 & 91\\
 Sydney Siege &  5 & 1 & 12 & 69\\
 Charlie Hebdo &  9 & 2 & 8 & 74\\
 Germanwings &  11 & 5 & 15 & 71\\
 Marina Joyce &  5 & 30 & 10 & 110\\
 Hillary's Illness &  39 & 27 & 24 & 297\\
\hline
 \textbf{Total} & 94 & 71 & 106 & 778\\
\hline
\end{tabular}
\end{center}
\caption{\label{data-distribution} Semeval-2017 Task 8 (A) dataset distribution.}
\end{table}

{\bf Participants} Eight teams participated in the task. 
The best performing system was developed by Turing (78.4 in accuracy). 
ECNU, MamaEdha, UWaterloo, and DFKI-DKT utilized ensemble classifier.
Some systems also used deep learning techniques, including Turing, IKM, and MamaEdha. Meanwhile, NileTRMG and IITP used classical classifier (SVM) to build their systems. Most of the participants exploited word embedding to construct their feature space, beside the Twitter domain features.

\section{Proposed Method}

We developed a new model by exploiting
several stylistic and structural features characterizing Twitter language. In addition, we propose to utilize conversational-based features by exploiting the peculiar tree structure of the dataset. We also explored the use of affective based feature by extracting information from several affective resources including dialogue-act inspired features. 

\subsection{Structural Features} 
They were designed taking into account several Twitter data characteristics, and then selecting the most relevant features to improve the classification performance.
The set of structural features that we used is listed below.
\begin{itemize}
\item[]{\bf Retweet Count}: The number of retweet of each tweet.
\item[]{\bf Question Mark}: presence of question mark "?"; binary value (0 and 1).
\item[]{\bf Question Mark Count}: 
number of question marks present in the tweet.
\item[]{\bf Hashtag Presence}: this feature has a binary value 0 (if there is no hashtag in the tweet) or 1 (if there is at least one hashtag in the tweet). 
\item[]{\bf Text Length}: number of characters after removing Twitter markers such as hashtags, mentions, and URLs.
\item[]{\bf URL Count}: number of URL links in the tweet.
\end{itemize}

\subsection{Conversation Based Features} These features are devoted to exploit the peculiar characteristics of the dataset, which have a tree structure reflecting the conversation thread\footnote{The implementation 
of these features is inspired from unpublished shared code~\cite{graf_david_2017_1133830}.}.
\begin{itemize}
\item[]{\bf Text Similarity to Source Tweet}: Jaccard Similarity of each tweet with its source tweet.
\item[]{\bf Text Similarity to Replied Tweet}: the degree of similarity between the tweet with the previous tweet in the thread (the tweet is a reply to that tweet).
\item[]{\bf Tweet Depth}: the depth value is obtained by counting the node from sources (roots) to each tweet in their hierarchy.
\end{itemize}

\subsection{Affective Based Features}
The idea to use affective features in the context of our task was inspired by recent works on fake news detection, focusing on emotional responses to true and false rumors  
\cite{vosoughi2018spread}, and by the work in \cite{novielli2013role} reflecting on the role of affect in dialogue acts \cite{novielli2013role}. Multi-faceted affective features have been already proven to be effective in some related tasks \cite{lai2016friends}, including the stance detection task proposed at SemEval-2016 (Task 6).

We used the following affective resources relying on different emotion models. 

\begin{itemize}
\item[]{\bf Emolex}: it contains 14,182 words associated with eight primary emotion based on the Plutchik model \cite{mohammad2013crowdsourcing,plutchik2001nature}.\item[]{\bf EmoSenticNet(EmoSN)}: it is an enriched version of SenticNet \cite{cambria2014senticnet} including 13,189 words labeled by six Ekman's basic emotion \cite{poria2013enhanced,ekman1992argument}.
\item[]{\bf Dictionary of Affect in Language (DAL)}: includes 8,742 English words labeled by three scores representing three dimensions: Pleasantness, Activation and Imagery \cite{whissell2009using}.
\item[]{\bf Affective Norms for English Words (ANEW)}: 
consists of 1,034 English words \cite{bradley1999affective} rated with ratings based on the 
Valence-Arousal-Dominance (VAD) 
model \cite{osgood_measurement_1957}.
\item[]{\bf Linguistic Inquiry and Word Count (LIWC)}: this psycholinguistic resource \cite{pennebaker2001linguistic} includes 4,500 words distributed into 64 emotional categories including positive (PosEMO) and negative (NegEMO). 
\end{itemize}

\subsection{Dialogue-Act Features}
We also included additional 11 categories from {\ bf LIWC}, which were already proven to be effective in dialogue-act task in previous work~\cite{novielli2013role}. Basically, these features are part of the affective feature group, but we present them separately because we are interested in exploring the contribution of such feature set separately. This feature set was obtained by selecting 4 communicative goals related to our classes in the stance task: {\bf agree-accept} (support), {\bf reject} (deny), {\bf info-request} (question), and {\bf opinion} (comment). The 11 LIWC categories include:
\begin{itemize}
\item[]{\bf Agree-accept:} Assent, Certain, Affect;
\item[]{\bf Reject:} Negate, Inhib;
\item[]{\bf Info-request:} You, Cause;
\item[]{\bf Opinion:} Future, Sad, Insight, Cogmech.
\end{itemize}

\section{Experiments, Evaluation and Analysis}

We used the RumourEval dataset from SemEval-2017 Task 8 described in Section \ref{sec:dataset}. 
We defined the rumour stance detection problem as a simple four-way classification task,
where every tweet in the dataset (source and direct or nested reply) should be classified into one among four classes: support, deny, query, and comment. We conducted a set of experiments in order to evaluate and analyze the effectiveness of our proposed feature set.\footnote{We built our system by using scikit-learn Python Library:~\url{http://scikit-learn.org/} }.

The results are summarized in Table~\ref{performance-comparison}, showing that our system outperforms all of the other systems in terms of accuracy. Our best result was obtained by a simple configuration with a support vector classifier with radial basis function (RBF) kernel.
Our model performed better than the best-performing systems in SemEval 2017 Task 8 Subtask A (Turing team, \cite{kochkina2017turing}), which exploited deep learning approach by using LTSM-Branch model. In addition, we also got a higher accuracy than the system described in \cite{aker2017simple}, which exploits a Random Forest classifier and word embeddings based features. 

We experimented with several classifiers, including Naive Bayes, Decision Trees, Support Vector Machine, and Random Forest, noting that SVM outperforms the other classifiers on this task. 
We explored the parameter space by tuning the SVM hyperparameters, namely the penalty parameter C, kernel type, and class weights (to deal with class imbalance). We tested several values for C (0.001, 0.01, 0.1, 1, 10, 100, and 1000), four different kernels (linear, RBF, polynomial, and sigmoid) and weighted the classes based on their distribution in the training data. The best result was obtained with C=1, RBF kernel, and without class weighting.

\begin{table}
\begin{center}
\begin{tabular}{ p{0.5cm}p{4cm}p{1.5cm}  }
 \hline
   \textbf{No.} & \textbf{Systems} & \textbf{Accuracy} \\
 \hline
 1. & Turing's System & 78.4 \\
 2. & Aker et al. System & 79.02 \\
 3. & Our System & \textbf{79.5} \\
 \hline
 & RumourEval Baseline & 74.1\\
 \hline
\end{tabular}
\end{center}
\caption{\label{performance-comparison} Results and comparison with state of the art}
\end{table}

An ablation test was conducted to explore the contribution of each feature set. Table~\ref{ablation-test} shows the result of our ablation test, by exploiting several feature sets on the same classifier (SVM with RBF kernel) \footnote{Source code is available on the GitHub platform:\\
\url{https://github.com/dadangewp/SemEval2017-RumourEval}}. This evaluation includes macro-averages of precision, recall and $F_1$-score as well as accuracy. We also presented the scores for each class in order to get a better understanding of our classifier's performance. 

\begin{table}
\begin{center}
\begin{tabular}{ p{2cm} p{0.75cm} p{0.75cm} p{0.75cm} p{0.75cm} }
 \hline
   & \textbf{S} & \textbf{D} & \textbf{Q} & \textbf{C}\\
 \hline
 \textbf{Support} &  \textbf{27} & 0 & 3 & 64\\
 \hline
 \textbf{Deny} &  2 & \textbf{0} & 1 & 68\\
 \hline
 \textbf{Query} & 0 & 0 & \textbf{50} & 56\\
 \hline
 \textbf{Comment} & 13 & 0 & 8 & \textbf{757}\\
 \hline
\end{tabular}
\end{center}
\caption{\label{confusion-matrix} Confusion Matrix}
\end{table}

\begin{table}
\begin{center}
\begin{tabular}{ p{2cm} p{0.75cm} p{0.75cm} p{0.75cm} p{0.75cm} }
 \hline
   & \textbf{S} & \textbf{D} & \textbf{Q} & \textbf{C}\\
 \hline
 \textbf{Support} &  \textbf{39} & 14 & 5 & 13\\
 \hline
 \textbf{Deny} &  8 & \textbf{28} & 5 & 30\\
 \hline
 \textbf{Query} & 2 & 3 & \textbf{62} & 4\\
 \hline
 \textbf{Comment} & 14 & 14 & 2 & \textbf{41}\\
 \hline
\end{tabular}
\end{center}
\caption{\label{confusion-matrix-balance} Confusion Matrix on Balanced Dataset}
\end{table}

\begin{table*}[!ht]
\small
\centering
\begin{tabular}{cl|cccc|cccc|cccc|cccc}
 \hline
\textbf{Ablation Test} & \multicolumn{4}{r}{\textbf{Overall}} & \multicolumn{4}{r}{\textbf{Support}} & \multicolumn{4}{r}{\textbf{Query}} & \multicolumn{4}{r}{\textbf{Comment}} \\
 \hline
  \textbf{Set} & \textbf{Features} & Acc & Prec & Rec & F1 & Acc & Prec & Rec & F1 & Acc & Prec & Rec & F1 & Acc & Prec & Rec & F1\\ 
\hline
\hline
A & Structural & 0.731 & 0.41 & 0.37 & 0.38 & 0.18 & \textbf{0.28} & 0.18 & 0.22 & 0.39 & 0.56 & 0.39 & 0.46 & 0.91 & 0.78 & 0.91 & 0.84 \\
B & Conversational & 0.767 & 0.42 & 0.31 & 0.33 & 0.29 & 0.93 & 0.29 & 0.44 & 0 & 0 & 0 & 0 & 1 & 0.76 & 1 & 0.87 \\
C & Affective & 0.742 & 0.19 & 0.25 & 0.21 & 0 & 0 & 0 & 0 & 0 & 0 & 0 & 0 & 1 & 0.74 & 1 & 0.85 \\
D & Dialogue-Act & 0.742 & 0.19 & 0.25 & 0.21 & 0 & 0 & 0 & 0 & 0 & 0 & 0 & 0 & 1 & 0.74 & 1 & 0.85 \\
E & A + B & 0.783 & 0.54 & 0.43 & 0.45 & 0.29 & \textbf{0.73} & 0.29 & 0.41 & 0.42 & 0.62 & 0.42 & \textbf{0.52} & 0.96 & 0.8 & 0.96 & 0.87 \\
F & A + C & 0.741 & 0.42 & 0.36 & 0.38 & 0.14 & 0.27 & 0.14 & 0.18 & 0.39 & 0.62 & 0.39 & 0.48 & 0.93 & 0.77 & 0.93 & 0.84 \\
G & A + D & 0.736 & 0.42 & 0.37 & 0.38 & 0.18 & 0.3 & 0.18 & 0.23 & 0.37 & 0.59 & 0.37 & 0.45 & 0.92 & 0.77 & 0.92 & 0.84 \\
H & E + C & 0.788 & 0.56 & 0.42 & 0.46 & 0.28 & \textbf{0.74} & 0.28 & 0.4 & 0.44 & 0.7 & 0.44 & 0.54 & 0.97 & 0.8 & 0.97 & 0.87 \\
I & E + D & 0.784 & 0.53 & 0.43 & 0.46 & 0.3 & \textbf{0.65} & 0.3 & 0.41 & 0.45 & 0.67 & 0.45 & 0.54 & 0.96 & 0.8 & 0.96 & 0.87 \\
J & F + D & 0.749 & 0.43 & 0.36 & 0.38 & 0.14 & 0.33 & 0.14 & 0.19 & 0.38 & 0.63 & 0.38 & 0.47 & 0.94 & 0.77 & 0.94 & 0.85 \\
K & All Features & \textbf{0.795}& 0.57 & 0.43 & 0.47 & 0.29 & \textbf{0.73} & 0.29 & 0.41 & 0.47 & 0.75 & 0.47 & \textbf{0.58} & 0.97 & 0.8 & 0.97 & 0.88 \\
\hline
\hline
\end{tabular}
\raggedright *deny class is not presented, since the score is always zero (0)
\vspace{0.2cm}
\caption{\label{ablation-test} Ablation test on several feature sets.}
\end{table*}

%

Using only conversational, affective, or dialogue-act features (without structural features) did not give a good classification result. Set B (conversational features only) was not able to detect the query and deny classes, while set C (affective features only) and D (dialogue-act features only) failed to catch the support, query, and deny classes. Conversational features were able to improve the classifier performance significantly, especially in detecting the support class. Sets E, H, I, and K which utilize conversational features induce an improvement on the prediction of the support class (roughly from 0.3 to 0.73 on precision). Meanwhile, the combination of affective and dialogue-act features was able to slightly improve the classification of the query class. The improvement can be seen from set E to set K where the $F_1$-score of query class increased from 0.52 to 0.58. Overall, the best result was obtained by the K set which encompasses all sets of features. It is worth to be noted that in our best configuration system, not all of affective and dialogue-act features were used in our feature vector. After several optimization steps, we found that some features were not improving the system's performance. Our final list of affective and dialogue-act based features includes: \textbf{DAL Activation}, \textbf{ANEW Dominance}, \textbf{Emolex Negative}, \textbf{Emolex Fear}, \textbf{LIWC Assent}, \textbf{LIWC Cause}, \textbf{LIWC Certain} and \textbf{LIWC Sad}. Therefore, we have only 17 columns of features in the best performing system covering structural, conversational, affective and dialogue-act features.

We conducted a further analysis of the classification result obtained by the best performing system (79.50 on accuracy). Table \ref{confusion-matrix} shows the confusion matrix of our result. On the one hand, the system is able to detect the comment tweets very well. However, this result is biased due to the number of comment data in the dataset. On the other hand, the system is failing to detect denying tweets, which were falsely classified into comments (68 out of 71)\footnote{A similar observation is reported by the best team at Semeval-2017 \cite{kochkina2017turing}.}. Meanwhile, approximately two thirds of supporting tweets and almost half of querying tweets were classified into the correct class by the system. 

In order to assess the impact of class imbalance on the learning, we performed an additional experiment with a balanced dataset using the best performing configuration. We took a subset of the instances equally distributed with respect to their class from the training set (330 instances for each class) and test set (71 instances for each class). As shown in Table~\ref{confusion-matrix-balance}, our classifier was able to correctly predict the underrepresented classes much better, although the overall accuracy is lower (59.9\%). The result of this analysis clearly indicates that class imbalance has a negative impact on the system performance. 
 
\subsection{\em Error analysis} 
We conducted a qualitative error analysis on the 215 misclassified in the test set, to shed some light on the issues and difficulties to be addressed in future work and to detect some notable error classes. 

\noindent
{\bf Denying by attacking the rumour's author.}
An interesting finding from the analysis of the Marina Joyce rumour data is that it contains a lot of denying tweets including insulting comments towards the author of the source tweet, like in the following cases:

\begin{quotation}
{\em Rumour:} Marina Joyce\\
{\em Misclassified tweets:}\\
(da1) stfu you toxic sludge\\
(da2) @sampepper u need rehab \includegraphics[height=1em]{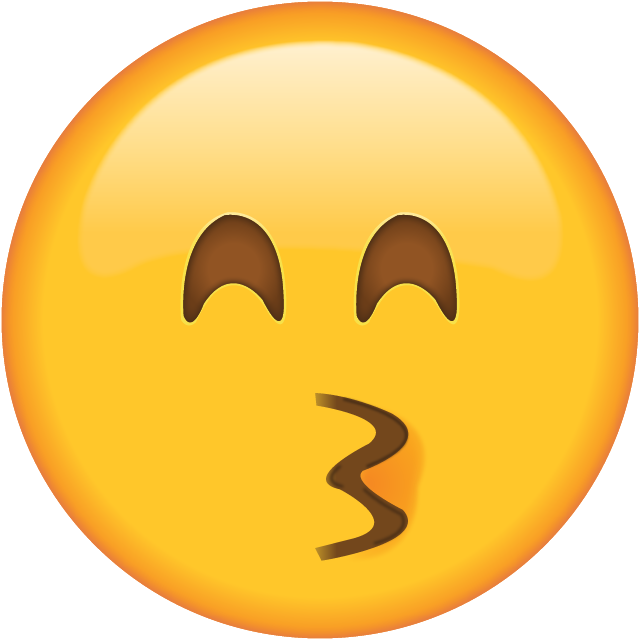}\\
{\em Misclassification type:} deny (gold) $\leadsto$ comment (prediction)\\
{\em Source tweet:}\\(s1) Anyone who knows Marina Joyce personally knows she has a serious drug addiction. she needs help, but in the form of rehab \#savemarinajoyce
\end{quotation}

\noindent
Tweets like (da1) and (da2) seem to be more inclined to show the respondent's personal hatred towards the s1-tweet's author than to deny the veracity of the rumour. In other words, they represent a peculiar form of denying the rumour, which is expressed by personal attack and by showing negative attitudes or hatred towards the rumour's author. This is different from denying by attacking the source tweet content, and it was difficult to comprehend for our system, that often misclassified such kind of tweets as comments.

\noindent
{\bf Noisy text, specific jargon, very short text.} In (da1) and (da2) (as in many tweets in the test set), we also observe the use of noisy text (abbreviations, misspellings, slang words and slurs, question statements without question mark, and so on) 
that our classifier struggles to handle . Moreover, especially in tweets from the Marina Joyce rumour's group, we found some very short tweets in the denying class that do not provide enough information, e.g. tweets like \emph{``shut up!"}, \emph{``delete"}, and \emph{``stop it. get some help"}.

\noindent
{\bf Argumentation context.} We also observed misclassification cases that seem to be related to a deeper capability of dealing with the argumentation context underlying the conversation thread.

\begin{quotation}
{\em Rumour:} Ferguson\\
{\em Misclassified tweet:}\\
(arg1)@QuadCityPat @AP I join you in this demand. Unconscionable.\\
{\em Misclassification type:} deny (gold) $\leadsto$ comment (prediction)\\
{\em Source tweet:}\\
(s2) @AP I demand you retract the lie that people in \#Ferguson were shouting ``kill the police", local reporting has refuted your ugly racism\\
\end{quotation}

\noindent
Here the misclassified tweet is a reply including an explicit expression of agreement with the author of the source tweet ({``\em I join you}''). Tweet (s2) is one of the rare cases of source tweets denying the rumor (source tweets in the RumourEval17 dataset are mostly supporting the rumor at issue). Our hypothesis is that it is difficult for a system to detect such kind of stance without a deeper comprehension of the argumentation context (e.g., if the author's stance is denying the rumor, and I agree with him, then I am denying the rumor as well). In general, we observed that when the source tweet is annotated by the {\em deny} label, most of denying replies of the thread include features typical of the support class (and vice versa), and this was a criticism.

\noindent
{\bf Mixed cases.} Furthermore, we found some borderline mixed cases in the gold standard annotation. See for instance the following case:

\begin{quotation}
{\em Rumour:} Ferguson\\
{\em Misclassified tweet: }\\
(mx1) @MichaelSkolnik @MediaLizzy Oh do tell where they keep track of "vigilante" stats.  That's interesting.\\
{\em Misclassification type:} query (gold) $\leadsto$ comment (prediction)\\
{\em Source tweet:} \\
(s3) Every 28 hours a black male is killed in the United States by police or vigilantes. \#Ferguson\\
\end{quotation}

\noindent
Tweet (mx1) is annotated with a {\em query} label rather than as a {\em comment} (our system prediction), but we can observe the presence of a comment ({\em ``That's interesting''}) after the request for clarification, so it seems to be a kind of mixed case, where both labels make sense. 

\noindent
{\bf Citation of the source's tweet.} We have noticed many misclassified cases of replying tweets with error pattern {\em support} (gold) $\leadsto$ {\em comment} (our prediction), where the text contains a literal citation of the source tweet, like in the following tweet: {\em THIS HAS TO END ``@MichaelSkolnik: Every 28 hours a black male is killed in the United States by police or vigilantes. \#Ferguson''} (the text enclosed in quotes is the source tweet). Such kind of mistakes could be maybe addressed by applying some pre-processing to the data, for instance by detecting the literal citation and replacing it with a marker. 

\noindent
{\bf Figurative language devices.}
Finally, the use of figurative language (e.g., sarcasm) is also an issue that should be considered for the future work. Let us consider for instance the following misclassified tweets:

\begin{quotation}
{\em Rumour:} Hillary's Illness\\
{\em Misclassified tweets:}\\
(fg1) @mitchellvii  True, after all she can open a pickle jar.\\
(fg2) @mitchellvii Also, except for having a 24/7 MD by her side giving her Valium injections, Hillary is in good health! https://t.co/GieNxwTXX7\\
(fg3) @mitchellvii @JoanieChesnutt At the very peak yes, almost time to go down a cliff and into the earth.\\
{\em Misclassification type:} support (gold) $\leadsto$ comment (prediction)\\
{\em Source tweet:}\\ (s4) Except for the coughing, fainting, apparent seizures and "short-circuits," Hillary is in the peak of health.
\end{quotation}
All misclassified tweets (fg1-fg3) from the {\em Hillary's illness} data are replies to a source tweet (s4), which is featured by sarcasm. In such replies authors support the rumor by echoing the sarcastic tone of the source tweet. Such more sophisticated cases, where the supportive attitude is expressed in an implicit way, were challenging for our classifier, and they were quite systematically misclassified as simple comments.

\section{Conclusion}

In this paper we proposed a new classification model for rumour stance classification. 
We designed a set of features including structural, conversation-based,  affective and dialogue-act based feature. Experiments on the SemEval-2017 Task 8 Subtask A dataset show that our system based on a limited set of well-engineered features outperforms the state-of-the-art systems in this task, without relying on the use of sophisticated deep learning approaches. Although achieving a very good result, several research challenges related to this task are left open. Class imbalance was recognized as one the main issues in this task. For instance, our system was struggling to detect the deny class in the original dataset distribution, but it performed much better in that respect when we balanced the distribution across the classes.

A re-run of the RumourEval shared task has been proposed at SemEval 2019\footnote{\url{http://alt.qcri.org/semeval2019/}} and it will be very interesting to participate to the new task with an evolution of the system here described. 

\section*{Acknowledgements} 

Endang Wahyu Pamungkas, Valerio Basile and Viviana Patti were partially funded by Progetto di Ateneo/CSP 2016 ({\em Immigrants, Hate and Prejudice in Social Media}, S1618\_L2\_BOSC\_01).

\bibliographystyle{ACM-Reference-Format}
\bibliography{sample-bibliography}

\end{document}